\documentclass[letterpaper]{article} 
\usepackage{aaai24}  
\usepackage{times}  
\usepackage{helvet}  
\usepackage{courier}  
\usepackage[hyphens]{url}  
\usepackage{graphicx} 
\usepackage{natbib}  
\usepackage{caption} 
\usepackage{algorithm}
\usepackage{algorithmic}
\usepackage{amsmath}
\usepackage{multirow}
\usepackage{booktabs}
\usepackage{amssymb}
\usepackage{pifont}

\usepackage{newfloat}
\usepackage{listings}
\DeclareCaptionStyle{ruled}{labelfont=normalfont,labelsep=colon,strut=off} 
\floatstyle{ruled}
\newfloat{listing}{tb}{lst}{}
\floatname{listing}{Listing}
\title{Enhance Sketch Recognition's Explainability via Semantic Component-Level Parsing}
\author {
	Guangming Zhu\textsuperscript{\rm 1,2,3},
	Siyuan Wang\textsuperscript{\rm 1},
	Tianci Wu\textsuperscript{\rm 1},
	Liang Zhang\textsuperscript{\rm 1,2,3,}\footnote{Corresponding Author}
}
\affiliations {
	\textsuperscript{\rm 1}School of Computer Science and Technology, Xidian University, China\\
	\textsuperscript{\rm 2}Key Laboratory of Smart Human-Computer Interaction and Wearable Technology of Shaanxi Province\\
	\textsuperscript{\rm 3}Xi'an Key Laboratory of Intelligent Software Engineering\\	
	gmzhu@xidian.edu.cn, siyuanwang@stu.xidian.edu.cn, 22031212495@stu.xidian.edu.cn, liangzhang@xidian.edu.cn
}

\usepackage{bibentry}

\begin{document}

\maketitle

\begin{abstract}
Free-hand sketches are appealing for humans as a universal tool to depict the visual world. Humans can recognize varied sketches of a category easily by identifying the concurrence and layout of the intrinsic semantic components of the category, since humans draw free-hand sketches based a common consensus that which types of semantic components constitute each sketch category. For example, an airplane should at least have a fuselage and wings. Based on this analysis, a semantic component-level memory module is constructed and embedded in the proposed structured sketch recognition network in this paper. The memory keys representing semantic components of each sketch category can be self-learned and enhance the recognition network's explainability. Our proposed networks can deal with different situations of sketch recognition, i.e., with or without semantic components labels of strokes. Experiments on the SPG and SketchIME datasets demonstrate the memory module's flexibility and the recognition network's explainability. The code and data are available at \url{https://github.com/GuangmingZhu/SketchESC}.
\end{abstract}

\section{Introduction}
Free-hand sketch is a universal tool to depict the visual world, and it is not bound by age, race, language, geography, or national boundaries. Sketch images are highly sparse, abstract and lack of background. Sketch can be regarded as an expression of the human brain's internal representation of the visual world \cite{xu2022deep}. Humans can recognize sketches and identify the intrinsic semantic components easily, even sketches of the same category drawn by different persons may be very different in appearance.  

Sketch can be represented as an image in the static pixel space, as a time series in the dynamic stroke coordinate space, or as a graph in the geometric graph space. This results in various Convolutional Neural Network (CNN), Recurrent Neural Network (RNN), and Graph Neural Network (GNN) based methods for sketch recognition \cite{zhang2019survey, xu2022deep}. These methods usually take image- or Scalable Vector Graphics (SVG)- format data as input, and predict the category label for a given sketch sample. However, there is lacking of work on interpreting the reason of giving such predictions.

Explainable artificial intelligence (XAI) has become a hot research topic to explain models' decision \cite{ramaswamy2020ablation, shitole2021one, garau2022interpretable}. Visualizing the activation maps of deep neural networks is widely used in computer vision. However, sketch images composed of stroke lines without textures, are different from natural images. This means that the existing XAI methods cannot be applied directly in the sketch research field. A first look at explainability for human sketches was achieved by SketchXAI using the counterfactual explanation \cite{qu2023sketchxai}. The stroke location inversion module in SketchXAI offers an explainability angle to sketch in terms of asking a network how well it can recover stroke locations of an unseen sketch. Liu et al. developed an image classifier explanation model using the counterfactual maps, in which the counterfactual map generator module is used to identify the critical structures for the specific category \cite{liu2023prediction}.

Counterfactual explanation (CE), as a post-hoc explainability method, aims to identify what are the minimal input changes for a model to make a different visual decision \cite{van2021interpretable}. SketchXAI \cite{qu2023sketchxai} used CE to relocate reshuffled strokes to construct a sketch given a category, while Liu et al. designed a counterfactual map generator to discover the stroke-level principal components for a specific category \cite{liu2023prediction}. The above two methods try to explain the question of ``\textit{why the sketch is classified as X}'' by providing positive and negative semantic explanation evidences. However, we believe that \textit{the concurrence and layout of the intrinsic semantic components of a category can be a crucial evidence to explain the question from another perspective}. For example, taking into consideration the common knowledge that an airplane should at least have a fuselage and wings, if a sketch is composed of strokes which can be semantically grouped into a fuselage and wings, it probably is an airplane. As to the analysis above, we propose to enhance sketch recognition's explainability via semantic component-level parsing.

Specifically, a Semantic Component-level Memory (SCM) module is constructed, whose memory keys represent the semantic components of different sketch categories. The SCM module is embedded in a Structured Sketch Recognition (SSR) network, and evolves the stroke features based on the similarity with the learnable features of memory keys. The fused stroke-level or component-level features are fed into a Transformer to achieve a high recognition performance under the supervision on segmentation (if available) or compositionality (i.e., which types of semantic components constitute each sketch category). For the dataset with the category labels and the semantic component labels of strokes, the supervision on the component-level parsing in the SCM module and on the semantic segmentation results of the Transformer can be used to achieve a precise and explainable recognition performance. For the dataset only with category labels, the supervision on the compositionality can be used in the proposed SSR network to enhance the recognition network's explainability. This flexibility makes the proposed SCM module and SSR network applicable on sketch recognition and segmentation tasks and achieve better and explainable performance.

The main contribution of this study can be summarized as follows.
\begin{itemize}
\item A semantic component-level memory module is constructed, which can learn and store memory keys representing semantic components, and do explainable parsing from strokes to components.
\item A structured sketch recognition network is proposed, which has hierarchical and explainable abilities, from stroke-level embedding, component-level parsing to sketch-level recognition.
\item The proposed network is explainable and flexibility to deal with the sketch recognition situations with or without semantic component labels of strokes, and can achieve remarkable performance on the public datasets.
\end{itemize}

\section{Related Work}
\subsection{Sketch Recognition}
Sketches are generally represented as pixel-level rasterized images or ordered sequences of point coordinates. Typically, CNNs \cite{yu2017sketch, prabhu2018distribution}, RNNs \cite{sarvadevabhatla2016enabling,ha2017neural}, or CNN-RNN architectures \cite{xu2018sketchmate, li2020sketch} were constructed for sketch recognition. Recently, the trend from Euclidean (CNN, RNN based) to topological analysis (GNN based) has emerged in sketch recognition. A sketch can also be represented as the sparsely connected graphs in the topological space. Therefore, GNN based models were proposed to model sketch's local and global topological stroke structures \cite{xu2021multigraph}. There is no consensus on which representation style is better than the other, as each has its own merits based on the application scenarios. Rasterized images ignore the sketching orders and are better for offline recognition. Sequence-based representation can be used to continuously predict the labels using accumulated sketch strokes online, and can be used in more interactive real-time applications. Graph-based representation is flexible to encode local and global geometric sketch structures, and can be used for sketch grouping or segmentation. However, no matter which representation style is used, visual explanation is rarely studied for sketch recognition.

\subsection{Visual Explanation}
Various activation map visualization techniques, such as the Grad-CAM series methods\cite{ramprasaath2017gradcam, aditya2018gradcam, omeiza2019smooth}, have been widely researched to interpret the classifier's decision-making rationale. These methods highlight the essential regions, but the explainability on sketch recognition is better to explore strokes' effects on recognition, therefore they are not suitable for sketch researches. Contrasted to these pixel-level methods, patch-level methods tried to use representative patches to explain the classifier's prediction \cite{chen2019looks, zhang2018interpreting, ge2021peek}. However, considering the surrounding or overlapping between strokes of a sketch in the spatial layout, patches can not always represent individual semantic components. Besides, explanation via visualization is hard to understand for non-expert users. 

Counterfactual explanation methods \cite{van2021interpretable, miller2019explanation} supplied alternative approaches to identify what are the minimal input changes for a model to make a different visual decision. SketchXAI\cite{qu2023sketchxai} used CE to relocate reshuffled strokes to construct a sketch given a category, while Liu et al. designed a counterfactual map generator to discover the stroke-level principal components for a specific category \cite{liu2023prediction}. These methods contribute the first exploration on sketch recognition's explainability in the stroke-level.

Humans draw free-hand sketches based a common consensus that which types of semantic components constitute each sketch category. Strokes of a sketch can be considered as abstract representation of the object's shape, component, or attributes. Therefore, since humans perceive the visual world by parsing objects' shape, components and attributes hierarchically and structurally, why cannot sketch recognition networks enhance their explainability by identifying the semantic components that strokes constitute. Alanize et al. constructed a Primitive-Matching Network (PMN) to learn interpretable abstracts of a sketch through simple primitives \cite{alaniz2022abstracting}. Zhu et al. proposed a simultaneous sketch recognition and segmentation (SketchRecSeg) network which parses the semantic components at the same time when recognizing a sketch \cite{zhu2023sketch}. However, PMN \cite{alaniz2022abstracting} only fulfills the matching between strokes and primitives. SketchRecSeg \cite{zhu2023sketch} uses a two-stream architecture, but its segmentation stream cannot enhance its recognition stream's explainability. In this paper, we construct a semantic component-level memory module and embed it in a structured recognition network to enhance sketch recognition's explainability. 

\section{Methodology}

\begin{figure}[t]
\centering
\includegraphics[width=\columnwidth]{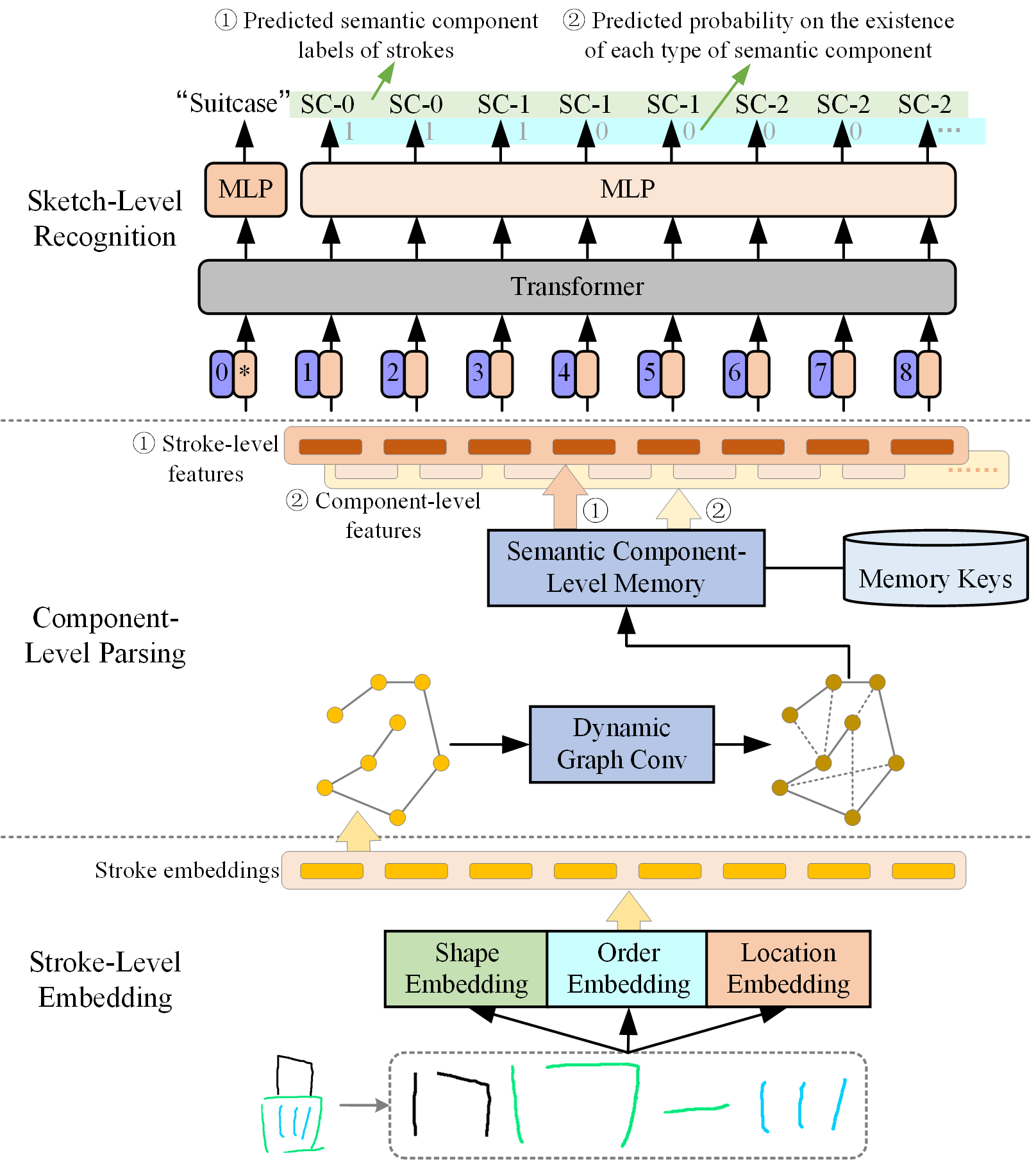} 
\caption{Overview of the proposed Structured Sketch Recognition network. The \ding{172} indicates the scenario that the Semantic Component-Level Memory module feeds the fused stroke-level features into Transformer for sketch recognition and segmentation. The \ding{173} indicates the scenario that the fused component-level features are fed into Transformer for sketch recognition and the probability prediction on the existence of each type of semantic component.}
\label{fig1}
\end{figure}

We aim to construct a Structured Sketch Recognition (SSR) network which does Stroke-Level Embedding on each stroke, implements Component-Level Parsing, and fulfills explainable Sketch-Level Recognition, as shown in Fig. \ref{fig1}. For the data with category labels and the semantic component labels of each stroke (i.e., the scenario \ding{172} in Fig. \ref{fig1}), \textit{sketches can be recognized and semantically segmented simultaneously}. For the data only with category labels and the prior knowledge about the intrinsic semantic components of each category  (i.e., the scenario \ding{173} in Fig. \ref{fig1}), \textit{sketches can be recognized with the auxiliary constraint that which types of semantic components constitute each sketch category}. Both two scenarios result in sketch recognition results with the auxiliary information about which types of semantic components constitute each sketch sample. This enhances the sketch recognition network's explainability.

\subsection{Stroke-Level Embedding}
Formally, each sketch can be represented as an ordered sequence of strokes, denoted as $\{s_1,s_2,\cdots,s_i,\cdots,s_N\}$. Stroke $s_i$ consists of $k$ points, $\{s_{i,1},s_{i,2},\cdots,s_{i,k-1},s_{i,k}\}$, and each point contains a two-dimensional coordinate value and a two-dimensional binary pen state \cite{ha2018neural}. Three descriptors are learned to identify three inherent properties of each sketch, i.e., shape $sh_i$, stroke order $o_i$ and location $l_i$, as in SketchXAI \cite{qu2023sketchxai}. The location of stroke $s_i$ is defined as the coordinate of the first point $s_{i,1}$. Specifically, a bidirectional Long Short-Term Memory (LSTM) is used for the shape embedding to extract shape information $sh_i$ of each stroke, a learnable embedding is used for the order embedding $o_i$, and one linear layer is used for the location embedding $l_i$. These three kinds of embeddings are summed as the stroke embeddings.

\subsection{Component-Level Parsing}
When a sketch is represented as a sparsely connected graph, the graph nodes generally denote the stroke points, as in SketchGNN \cite{yang2021sketchgnn} and MultiGraph Transformer \cite{xu2021multigraph}. In this study, stroke points have been aggregated in the stroke-level embedding stage, therefore a stroke-level graph $\mathcal{G}=(\mathcal{V}, \mathcal{E})$ can be constructed. $\mathcal{V}$ denotes the graph node set, in which each node represents a stroke. $\mathcal{E}$ denotes edges that connect adjacent strokes in sketching order.

\subsubsection{Dynamic Graph Convolution}
The above stroke-level embedding does not involve inter-stroke feature fusion. However, the semantic meaning of a stroke does not only depend on its shape and location, but also depends on the context of strokes. Inter-stroke fusion is necessary to learn which strokes constitute a semantic component. A two-layer dynamic graph convolution \cite{yang2021sketchgnn} unit is used in our network. The same graph convolution operation as in EdgeConv \cite{wang2019dynamic} is adopted. In order to enlarge the receptive field, $\mathcal{E}$ is updated layer-by-layer using the Dilated k-NN \cite{li2019deepgcns}. The motivation of updating $\mathcal{E}$ is to explore the feature fusion between strokes which belong to the same semantic component but are not adjacent in sketching order. The dilation ratios in the two layers are 1 and 2, respectively. A residual connection exists in each graph convolution layer to sum the input and output features. 

\subsubsection{Semantic Component-Level Memory}
Memory augmented neural networks utilize external memory to explicitly access the experiences \cite{khasahmadi2019memory}. A Semantic Component-level Memory (SCM) module can store the feature representation of semantic components, so that a similarity metric can be implemented to associate strokes with the semantic components to which they belong. In such case, strokes belonging to the same semantic component can be fused further to get the component-level features. Explainable similarity metrics ensure the explainability of the semantic component-level parsing, and the category classifier can do explainable inference, i.e., ``\textit{The sketch is recognized as X because it is composed of the semantic components which constitute X}".

Specifically, the SCM module consists of a multi-head array of memory keys. Each semantic component is represented by a multi-head key in SCM. Given the stroke feature $q_i$ outputted by the dynamic graph convolution module, we use Eq. (\ref{kernel}) as a kernel to measure the normalized similarity between the stroke feature $q_i$ and the key $k_j$ of SCM as\footnote{The head index of multi-head keys is omitted for simplicity }:

\begin{equation}
\label{kernel}
C_{i,j}=\frac{(\varepsilon +\Vert q_i-k_j\Vert^2/\tau)^{-\frac{\tau+1}{2}}}{\sum_{j'}{(\varepsilon+\Vert q_i-k_{j'} \Vert^2/\tau)^{-\frac{\tau+1}{2}}}},
\end{equation} 
where $C_{i,j}$ is the normalized score between the stroke feature $q_i$ and the memory key $k_j$ (representing the $j$-th type of semantic component), $\tau$ is the degree of freedom, and $\varepsilon$ is a bias value which is much smaller than the average of $\Vert q_i-k_j\Vert^2/\tau$. \textit{Memory keys are learnable parameters and learned automatically during the network training process.}

A Max-pooling operation is implemented to select the most similar head from the multi-head key of each semantic component for each stroke. For simplicity, we use $C_{i,j}$ to represent the similarity between the stroke $q_i$ and its most similar head key $k_j$ of the $j$-th type of semantic component in the following description, and use $\mathbf{K} \in \mathcal{R}^{K\times d}$ to represent the set of the most similar head key of each semantic component for one stroke, where $K$ is the component type count. A Softmax operation is further implemented along the $j$-dimension of $\{C_{i,j}\}$ to obtain the normalized assignment matrix $\mathbf{C} \in \mathcal{R}^{N\times K}$, where $N$ is the stroke count. 

\textbf{\textit{Feature Fusion.}} Two feature fusion strategies are designed. One is the stroke-level feature fusion, i.e., \textit{enhancing the stroke features by memory keys}, denoted as

\begin{equation}
\mathbf{F}_s\in\mathcal{R}^{N\times d} =(1-max_j(\mathbf{C}))\circ\mathbf{Q}+max_j(\mathbf{C})\circ\mathbf{C}*\mathbf{K}.
\label{fs}
\end{equation} 

\noindent The enhanced features $\mathbf{F}_s$ are further fed into Transformer for sketch recognition and segmentation. The other is the component-level feature fusion, i.e., \textit{generating component features by fusing stroke features and memory keys},  

\begin{equation}
\mathbf{F}_c\in\mathcal{R}^{K\times d} =(1-max_j(\mathbf{C}))\circ\mathbf{C}^{\top}*\mathbf{Q}+max_j(\mathbf{C})\circ\mathbf{K}.
\label{fc}
\end{equation} 

\noindent The component features $\mathbf{F}_c$ can be fed into Transformer for sketch recognition along with the prediction on the existence of each type of semantic component. In Eqs. (2) and (3), $\mathbf{Q}\in \mathcal{R}^{N\times d}$ is the stroke features outputted by the dynamic graph convolution module, $\circ$ is the broadcasting multiply operation, and $*$ is the matrix multiplication operation.  The balance ratio $max_j(\mathbf{C})$ means that if a stroke can be assigned to a semantic component with a high confidence, the key feature of the semantic component is more representative and better used for sketch recognition.

\textbf{\textit{Supervision on SCM.}} The keys in MemGNN are learned without extra supervision \cite{khasahmadi2019memory}. We believe that it is better to ensure keys' distinguishability, since keys represent different types of semantic components. Therefore, a linear classifier and a \textit{Cross-Entropy} (\textit{CE}) loss are implemented in the SCM module as

\begin{equation}
\mathcal{L}1=CE(f_{w1}(k_j),j).
\end{equation}

\noindent Besides, if the semantic component label of each stroke is available, a supervision on the assignment matrix $\mathbf{C}$ can be implemented by a \textit{balanced Binary Cross-Entropy} (\textit{bBCE}) loss as 

\begin{equation}
\begin{aligned}
\mathcal{L}2&=bBCE(\mathbf{C},\mathbf{C}^{gt})\\
&=\gamma_n\sum{C^{gt}_{i,j}C_{i,j}} + \gamma_p\sum{(1-C^{gt}_{i,j})(1-C_{i,j})}, 
\end{aligned}
\end{equation}
\noindent where the $(i,j)$-th value $C^{gt}_{i,j}$ in $\mathbf{C}^{gt} \in \mathcal{R}^{N\times K}$ is 1 when the $i$-th stroke belongs to the $j$-th type of semantic component, otherwise the value is 0. The $\gamma_n$ and $\gamma_p$ denote the ratio of 0 and 1 in $\mathbf{C}^{gt}$, respectively. The balance ratio $\gamma_n$ and $\gamma_p$ prevent the $\mathbf{C}$ from being learned as all-zero, since only one of $K$ elements in each row of $\mathbf{C}^{gt}$ is 1.

\subsection{Sketch-Level Recognition}
The Transformer architecture in ViT \cite{dosovitskiy2020image} is used for sketch-level recognition. When taking the fused stroke-level features $\mathbf{F}_s$ as input, the Transformer outputs the category label and the semantic component label of each stroke. The classification ($\mathcal{L}4$) and stroke-level semantic segmentation ($\mathcal{L}5$) losses can be denoted as

\begin{equation}
\label{L31}
\mathcal{L}3=
\underbrace{CE(f_{w2}(\mathbf{F}_s), y_c)}_{\mathcal{L}4}
+ \lambda_s
\underbrace{CE(f_{w3}(\mathbf{F}_s), \mathbf{y}_s)}_{\mathcal{L}5},
\end{equation}

\noindent where $y_c$ is the ground-truth category label and $\mathbf{y}_s$ is the ground-truth semantic component label of strokes.

When taking the fused component-level features $\mathbf{F}_c$ as input, the Transformer outputs the category label and the prediction probability on the existence of each semantic component in the sketch sample. The classification ($\mathcal{L}4$) and compositionality prediction ($\mathcal{L}6$) losses can be denoted as

\begin{equation}
\label{L32}
\mathcal{L}3=
\underbrace{CE(f_{w2}(\mathbf{F}_s), y_c)}_{\mathcal{L}4}
+ \lambda_c
\underbrace{bBCE(f_{w4}(\mathbf{F}_c), \mathbf{y}_e)}_{\mathcal{L}6},
\end{equation}

\noindent where $\mathbf{y}_e$ indicates the existence or not of each type of semantic component. $\mathbf{y}_e^j=1$ when the sketch sample should contain the $j$-th type of semantic component, otherwise $\mathbf{y}_e^j=0$. The component-level features $\mathbf{F}_c$ has fixed $K$ feature vectors, no matter how many strokes are contained in the sketch sample. Therefore, $\mathbf{y}_e$ is sparse, and a balanced binary cross-entropy loss is used (denoted as $\mathcal{L}6$).

\subsection{Losses}

The overall loss can be calculated as

\begin{equation}
\mathcal{L}=\lambda_1\mathcal{L}1+\lambda_2\mathcal{L}2+\mathcal{L}3.
\label{allloss}
\end{equation}

$\mathcal{L}1$ ensures the distinguishability of the memory keys in SCM, and it does not need the semantic component labels of keys or strokes. 

$\mathcal{L}2$ works only when the dataset has the semantic component labels of strokes. If not, the memory keys are learned without the direct supervision on the assignment matrix $\mathbf{C}$. 

$\mathcal{L}3$ in Eq. (\ref{L31}) works for sketch recognition and segmentation. If the semantic component labels of strokes are unavailable but the prior information about which types of semantic components constitute each sketch category is known, $\mathcal{L}3$ in Eq. (\ref{L32}) can help the Transformer achieve a better and explainable recognition performance.

\section{Experiments}
\subsection{Datasets}
The SPG dataset \cite{li2018universal} and SketchIME dataset \cite{zhu2023sketch} are used to verify the advantages of the proposed network. SPG was originally constructed for sketch perceptual grouping, and the same 20 categories as in SketchGNN \cite{yang2021sketchgnn} are used for evaluation. An average of 600 samples per category are used for training, while 100 samples for testing. Total 87 types of semantic components are defined according to the original labels in SPG to support our researches. SketchIME is a systematic dataset comprising 374 specialized sketch categories. Total 139 types of semantic components are defined. This study selects 56K samples which have category labels and semantic component labels of strokes from the released 209K samples. An average of 100 samples per sketch category are used for training, while 50 samples for testing.

\subsection{Evaluation Metrics } 
The Top-1 accuracy (Acc@1) is used as the evaluation metric for sketch recognition. SketchSegNet \cite{wu2018sketchsegnet} and SketchGNN \cite{yang2021sketchgnn} used point-based accuracy and component-based accuracy for sketch segmentation. Since the proposed SSR network does predictions on strokes for semantic segmentation directly, only the component-based accuracy (C-Metric) which indicates the percentage of the correctly predicted strokes is used as the evaluation metric for segmentation.

\begin{table}[t]
\centering
\resizebox{\columnwidth}{!}{
\begin{tabular}{llccccc}
\toprule
Available Labels & SCMFeat & w/ $\mathcal{L}2$ & w/ $\mathcal{L}5$ & w/ $\mathcal{L}6$ & Acc@1 & C-Metric \\
\hline
\multirow{3}{*}{C-Labels Only} & $\mathbf{F}_s$ as Eq. (\ref{fs}) &  &  &  & 88.48 & - \\
& $\mathbf{F}_s=\mathbf{C}*\mathbf{K}$ &  &  &  & 91.41 & - \\
& $\mathbf{F}_s=\mathbf{Q} $ &  &  &  & 91.01 & - \\
\hline
\multirow{4}{*}{C-Labels and Prior Info} & $\mathbf{F}_c$ as Eq. (\ref{fc}) &  &  &  & 92.02 & - \\
& $\mathbf{F}_c=\mathbf{C}^{\top}*\mathbf{Q}$ &  &  &  & 90.71 & - \\
& $\mathbf{F}_c$ as Eq. (\ref{fc}) &  &  & \checkmark & 94.04 & - \\
& $\mathbf{F}_c=\mathbf{C}^{\top}*\mathbf{Q}$ &  &  & \checkmark & 94.55 & - \\
\hline
\multirow{3}{*}{C-Labels and SC-Labels} & $\mathbf{F}_s$ as Eq. (\ref{fs}) & \checkmark & \checkmark &  & 95.81 & \textbf{90.12} \\
& $\mathbf{F}_s=\mathbf{C}*\mathbf{K}$ & \checkmark & \checkmark &  & \underline{96.62} & \underline{89.69} \\
& $\mathbf{F}_s=\mathbf{Q} $ & \checkmark & \checkmark &  & \textbf{96.67} & 89.42 \\
\bottomrule
\end{tabular}}
\caption{The sketch recognition and segmentation performance on the SPG dataset under different configurations. ``SCMFeat'' denotes which kinds of features are fed into Transformer by the SCM module. ``C-Labels'' means the category labels, and ``SC-Labels'' denotes the semantic component labels of strokes. ``Prior Info'' represents the prior information about which types of semantic components constitute each sketch category. The losses $\mathcal{L}1$ and $\mathcal{L}4$ are always used, but $\mathcal{L}2$, $\mathcal{L}5$ and $\mathcal{L}6$ may not be used when different labels are available.}
\label{table1}
\end{table}

\subsection{Network Details}
In the stroke-level embedding module, a bidirectional LSTM layer takes a sequence of 4-dimensional stroke points as input and outputs a 768-dimensional shape embedding, a linear layer transforms a two-dimensional coordinate into a 768-dimensional location embedding, and the 768-dimensional order embedding is learned by PyTorch's nn.Embedding function. In the dynamic graph convolution module, the number of neurons in each convolution layer is all 768. The same Transformer as ViT-Base \cite{dosovitskiy2020image} is used for sketch-level recognition.

\subsection{Training Details}
The learning rate is initialized to $3\times 10^{-4}$ with a batch size of 128. The Adam optimizer is used. Total 200 epochs are implemented for each training. The $\tau$ in Eq. (\ref{kernel}) is set to 1. The $\lambda_1$ and $\lambda_2$ in Eq. (\ref{allloss}) are set to 1 and 20, respectively. The $\lambda_s$ in Eq. (\ref{L31}) and the $\lambda_c$ in Eq. (\ref{L32}) are set to 10 empirically. The SSR network is trained from scratch, except the Transformer module initialized with the pretrained ViT-Base model from HuggingFace \footnote{https://huggingface.co/}. Our network is implemented by Pytorch and trained on a single NVIDIA GTX 3090.

\subsection{Ablation Study}

As aforementioned, the proposed SCM module and SSR network can deal with different cases with or without semantic component labels of strokes. As illustrated in Table \ref{table1}, three cases which use different features and losses are evaluated.

\textbf{Firstly}, when the category labels and semantic component labels of strokes are available, the supervision on the assignment matrix $\mathbf{C}$ (i.e., ``w/ $\mathcal{L}2$'') and on the prediction of the semantic component labels of each stroke (i.e., ``w/ $\mathcal{L}5$'') can be used. The multi-rows of the case ``C-Labels and SC-Labels'' in Table \ref{table1} illustrate the evaluation results. No matter which kinds features outputted by the SCM module are fed into Transformer for recognition and segmentation, excellent performances are achieved compared with the cases without semantic component labels of strokes. ``$\mathbf{F}_s=\mathbf{Q}$'' means that the stroke features learned by the dynamic graph convolution module are fed into Transformer, while ``$\mathbf{F}_c=\mathbf{C}*\mathbf{K}$'' means that the transformed memory keys are fed into Transformer. Both the two cases have achieve comparable performance. This means the learned memory keys can represent the semantic components effectively, although the memory keys are not calculated from the stroke features directly. ``$\mathbf{F}_s=\mathbf{Q}$'' does not mean the SCM module is excluded from the learning process, $\mathbf{Q}$ is still partially updated according to the gradient propagation from the supervision on the assignment matrix $\mathbf{C}$.

\textbf{Secondly}, when the semantic component labels of strokes are unavailable but the prior information about which types of semantic components constitute each sketch category is known, the prior information still can be used to enhance recognition's performance. The multi-rows of the case ``C-Labels and Prior Info'' in Table \ref{table1} illustrate the evaluation results. In such case, the supervision on the existence of each type of semantic component given a sketch can be used (i.e., ``w/ $\mathcal{L}6$''). The stroke features cannot be fed into Transformer directly, since Transformer cannot be supervised on the semantic component prediction for strokes. Therefore, the component-level features transformed from the stroke features based on the assignment matrix $\mathbf{C}$ are fed into Transformer. The four rows show that using the supervision can improve recognition performance significantly (92.02\% vs. 94.04\% and 90.71\% vs. 94.55\%). It also makes the recognition explainable, since Transformer can tell which types of semantic components are contained in each sketch sample, although it does not know to which type of semantic component each stroke belong. 

\textbf{Thirdly}, when neither the semantic component labels of strokes nor the prior information are available, the proposed network can still be used as a typical recognition network, as illustrated in the case of ``C-Labels Only'' in Table \ref{table1}. Both the fused stroke-level features $\mathbf{F}_s$ (i.e., see the three rows of the case of ``C-Labels Only'') and the fused component-level features $\mathbf{F}_s$ (i.e., see the top two rows of the case of ``C-Labels and Prior Info'') can be fed into Transformer which only does the prediction of category labels. It is unsurprising that the performances are not so good, since the memory keys are hard to learn without any extra supervision.

\textbf{In conclusion}, it is expected that more supervision can result in better performance, and the proposed method gives a flexible and explainable architecture to deal with sketch recognition with different auxiliary information.

\subsection{Comparison with State-of-The-Art}

Table \ref{table2} gives the comparison results with the state-of-the-art methods on the SPG dataset. The proposed SSR network outperforms all the methods except SketchRecSeg \cite{zhu2023sketch}. This is because SketchRecSeg is a two-stream network which takes both image- and SVG-format data as input, but the proposed SSR network only uses the SVG-format data. Besides, SketchSegNet \cite{wu2018sketchsegnet}, SketchGNN \cite{yang2021sketchgnn} and SketchRecSeg \cite{zhu2023sketch} all construct stroke point-level graphs and predict point-level segmentation labels. The proposed SSR network uses a hierarchical and structural architecture, stroke-level graphs are constructed and stroke-level predictions are performed. Furthermore, the proposed SSR network can fulfill simultaneous recognition and segmentation with the one-stream architecture, while SketchRecSeg \cite{zhu2023sketch} employs a two-stream architecture for recognition and segmentation, respectively. These factors demonstrate the superiority of the proposed SSR network.

Table \ref{table3} gives the comparison results on the selected SketchIME dataset which has 374 categories and 139 types of semantic components. The proposed SSR network still obtains the superior performance. This exactly demonstrates the applicability on large-scale datasets.

\begin{table}[t]
\centering
\resizebox{\columnwidth}{!}{
\begin{tabular}{lcc}
\toprule
Networks &  Acc@1 & C-Metric \\
\hline
ViT \cite{dosovitskiy2020image} & 76.21 & - \\
BiGRU \cite{chung2014empirical} & 79.10 & - \\
ResNet18 \cite{xu2022deep} & 80.66 & - \\
MGT \cite{xu2021multigraph} & 91.05 & - \\
\hline
SketchSegNet \cite{wu2018sketchsegnet} & - & 45.46 \\
SketchGNN \cite{yang2021sketchgnn} & - & 87.86 \\
\hline
SketchRecSeg \cite{zhu2023sketch} & \textbf{97.47} & \textbf{91.65} \\
\hline
SSR($\mathbf{F}_s$ as Eq. (\ref{fs})) & 95.81 & \underline{90.12} \\
SSR($\mathbf{F}_s=\mathbf{C}*\mathbf{K}$) & 96.62 & 89.69 \\
SSR($\mathbf{F}_s=\mathbf{Q} $) & \underline{96.67} & 89.42 \\
\bottomrule
\end{tabular}}
\caption{Comparison with state-of-the-art methods on the SPG dataset. The proposed SSR network using all the losses in Eq. (\ref{L31}) and Eq. (\ref{allloss}). }
\label{table2}
\end{table}

\begin{table}[t]
\centering
\resizebox{\columnwidth}{!}{
\begin{tabular}{lcc}
\toprule
Networks &  Acc@1 & C-Metric \\
\hline
ViT \cite{dosovitskiy2020image} & 22.02 & - \\
ResNet18 \cite{xu2022deep}  & 89.01 & - \\
MGT \cite{xu2021multigraph} & 70.31 & - \\
\hline
SketchSegNet \cite{wu2018sketchsegnet} & - & 61.78 \\
SketchGNN \cite{yang2021sketchgnn} & - & 94.01 \\
\hline
SSR($\mathbf{F}_s$ as Eq. (\ref{fs})) & \underline{89.88} & \underline{94.59} \\
SSR($\mathbf{F}_s=\mathbf{C}*\mathbf{K}$) & 87.92 & 94.43 \\
SSR($\mathbf{F}_s=\mathbf{Q} $) & \textbf{91.48} & \textbf{94.91} \\
\bottomrule
\end{tabular}}
\caption{Comparison with state-of-the-art methods on the SketchIME dataset. The proposed SSR network using all the losses in Eq. (\ref{L31}) and Eq. (\ref{allloss}). }
\label{table3}
\end{table}

\subsection{Visualization}

\begin{figure*}[t]
\centering
\includegraphics[width=\textwidth]{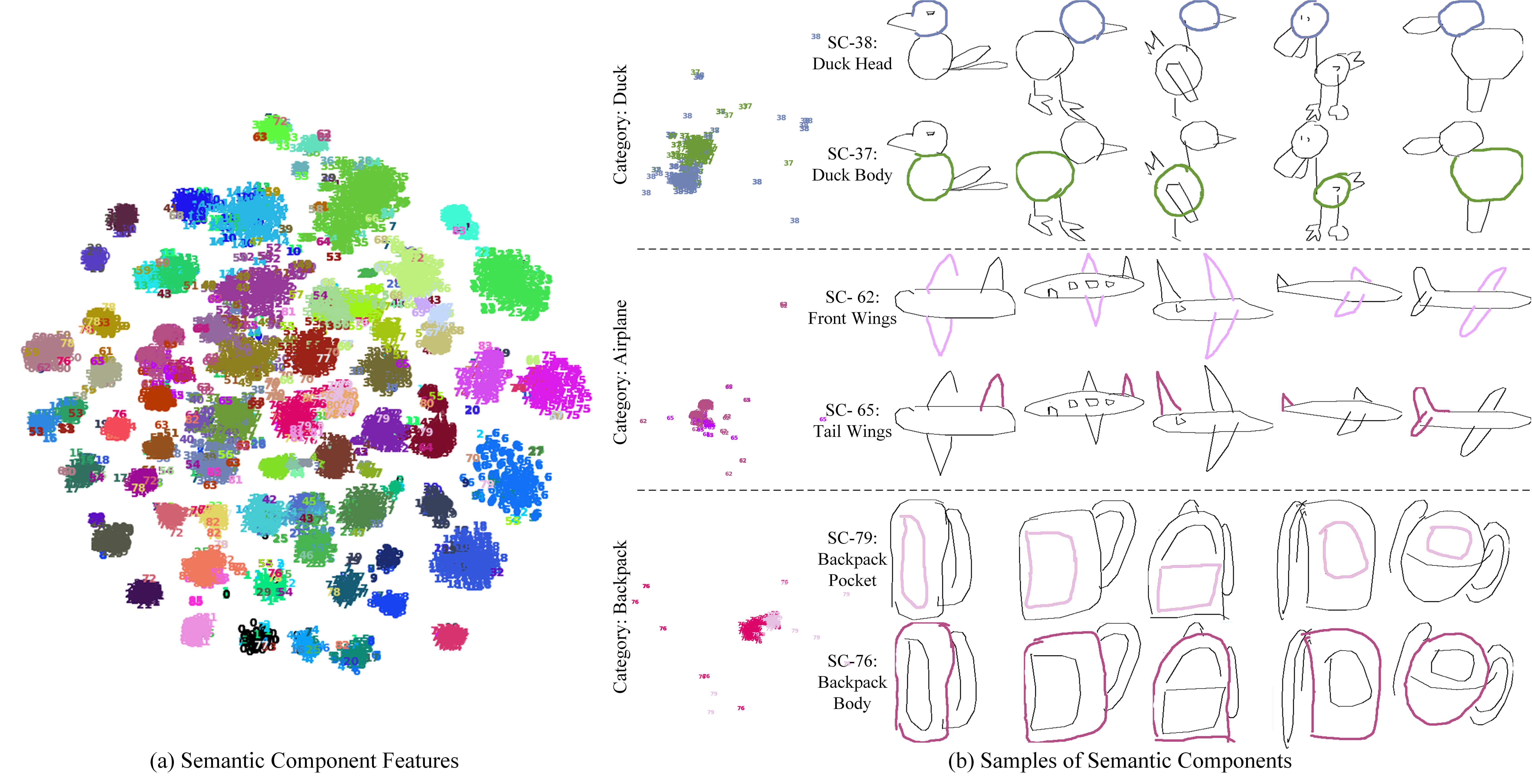} 
\caption{Visualization of semantic component features using t-SNE and some sketch samples. Fig. 2(a) shows the feature visualization of the 87 types of semantic components in SPG. Fig. 2(b) shows some sketch examples whose parts of semantic components are indistinguishable in the feature space, but our SSR network can do recognition and segmentation correctly.}
\label{fig2}
\end{figure*}

Figure \ref{fig2} gives the visualization of semantic component features using t-SNE \cite{van2008visualizing} and some sketch samples. It can be concluded from the feature visualization in Fig. \ref{fig2}(a) that,  the classification supervision on the memory keys of SCM ensures the distinguishability of the keys, and the SCM module further enhances the distinguishability of the strokes in the feature space. The memory mechanism can store the features of semantic components using multi-head arrays, and outperforms the mechanisms using classifiers to recognize the strokes' label directly or using Conditional Random Field (CRF) based methods \cite{yuan2020structpool} to learn strokes' clustering relationship. Figure \ref{fig2}(b) gives some sketch examples which have indistinguishable semantic components in the stroke feature space but are recognizable when considering the concurrence and layout of the semantic components. The proposed SSR network uses the SCM module to evolve stroke features in an explainable way, and uses Transformer to recognize the category label and the semantic component labels of strokes (or probabilities on the existence of each type of semantic component). This ensures the explainability of sketch recognition via semantic component-level parsing.

Figure 3 displays the recognition and segmentation results of some wrongly recognized sketch samples. It can be seen from Fig. 3 that these sketches are wrongly recognized because their strokes are wrongly resolved. Eqs. (2) and (3) show that the features fed into Transformer by the SCM module are calculated based on the stroke features outputted by the dynamic graph convolution module in an explainable way. Therefore, Transformer's prediction can be mapped into original strokes. This exactly demonstrates the explainability of our sketch recognition network.

\begin{figure}
\centering
\includegraphics[width=\columnwidth]{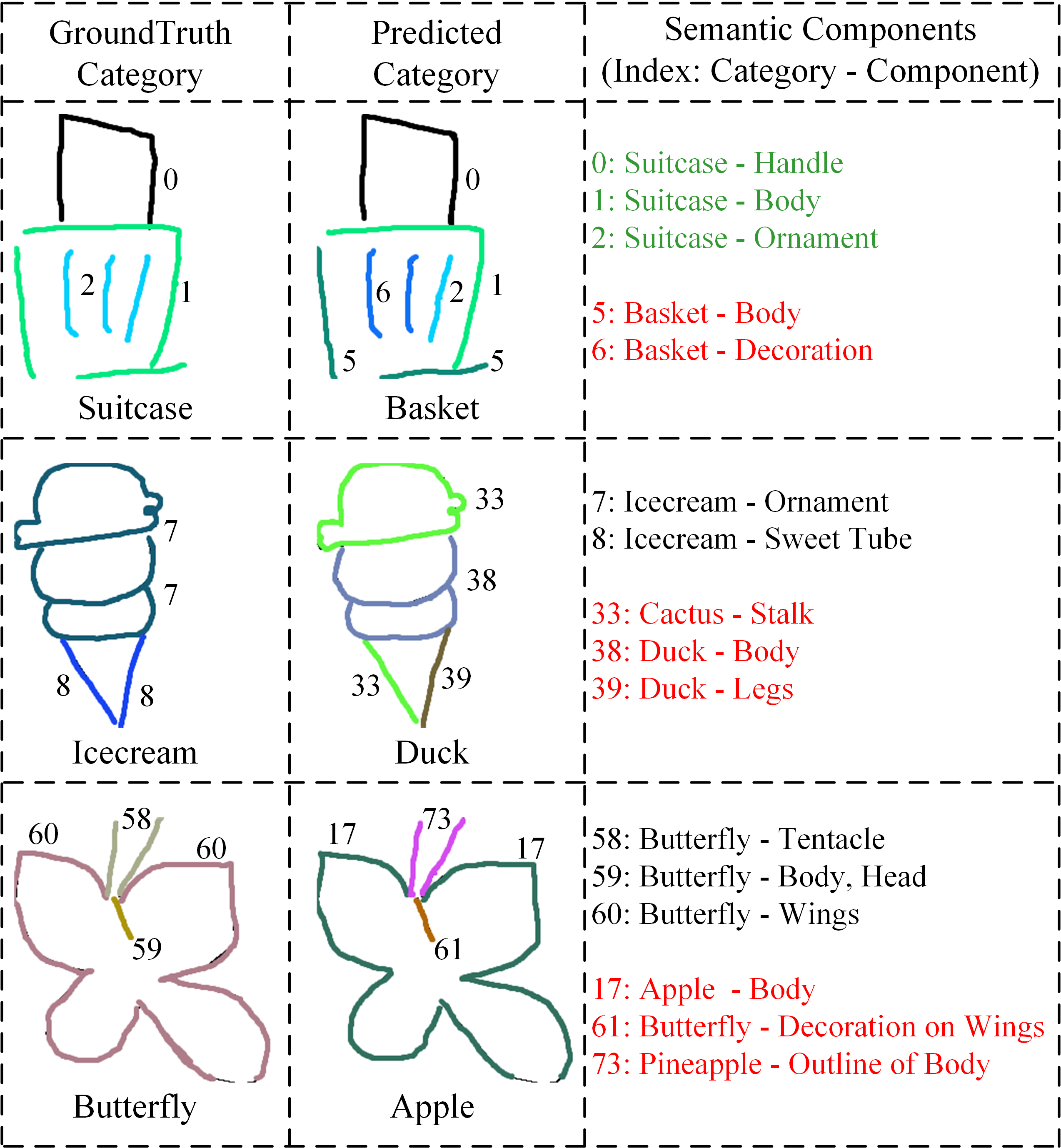} 
\caption{Examples of wrongly-recognized sketches. The numbers around the strokes are the groundtruth or predicted type indexes of semantic components.}
\label{fig3}
\end{figure}

\section{Discussion}
Activation map visualization techniques are not suitable for sketch recognition's explainability in the stroke-level. Counterfactual explanation based methods supply an alternative way, but SketchXAI \cite{qu2023sketchxai} only uses CE to explore the deserved layout of strokes of a sketch, and Liu et al. use CE to discover the stroke-level principal components for a specific category \cite{liu2023prediction}. They just partially answer the question of ``\textit{why the sketch is classified as X}''. This study answers the question from a perspective of semantic component-level parsing. Humans generally describe an object using sentences about its components and attributes. \textit{If a consensus can be reached that a sketch is represented structurally by some types of semantic components and their layout, we can easily find the superiority of our proposed network because the stroke-level embedding module can encode the layout of strokes, and the SCM and Transformer modules have abilities to resolve the semantic components.} The proposed SSR network gains sketch recognition's explainability in a more understandable and explainable way.

\section{Conclusion}
Deep learning based sketch recognition networks have achieved remarkable performance that even beats humans. However, humans can explain ``\textit{why the sketch is classified as X}'' easily, while sketch recognition networks are lacking of interpretable reasons of giving such predictions. This study tries to explore sketch recognition's explainability via semantic component-level parsing. A semantic component-level memory module is constructed, which can learn and store features of semantic components in multi-head arrays, and parse the strokes in the component-level. A structured sketch recognition network is proposed and equipped with the memory module. The network gives the explanation ``\textit{The sketch is recognized as X because it is composed of the semantic components which constitute X}''.

\section{Acknowledgments}
This work is partially supported by the National Natural Science Foundation of China under Grant No.62073252. and No.62072358.

\clearpage

\bibliography{aaai24}

\begin{thebibliography}{35}
\providecommand{\natexlab}[1]{#1}

\bibitem[{Alaniz et~al.(2022)Alaniz, Mancini, Dutta, Marcos, and
  Akata}]{alaniz2022abstracting}
Alaniz, S.; Mancini, M.; Dutta, A.; Marcos, D.; and Akata, Z. 2022.
\newblock Abstracting sketches through simple primitives.
\newblock In \emph{ECCV}, 396--412.

\bibitem[{Chattopadhay et~al.(2018)Chattopadhay, Sarkar, Howlader, and
  Balasubramanian}]{aditya2018gradcam}
Chattopadhay, A.; Sarkar, A.; Howlader, P.; and Balasubramanian, V.~N. 2018.
\newblock Grad-CAM++: Generalized Gradient-Based Visual Explanations for Deep
  Convolutional Networks.
\newblock In \emph{WACV}, 839--847.

\bibitem[{Chen et~al.(2019)Chen, Li, Tao, Barnett, Rudin, and
  Su}]{chen2019looks}
Chen, C.; Li, O.; Tao, D.; Barnett, A.; Rudin, C.; and Su, J.~K. 2019.
\newblock This looks like that: deep learning for interpretable image
  recognition.
\newblock volume~32.

\bibitem[{Chung et~al.(2014)Chung, Gulcehre, Cho, and
  Bengio}]{chung2014empirical}
Chung, J.; Gulcehre, C.; Cho, K.; and Bengio, Y. 2014.
\newblock Empirical evaluation of gated recurrent neural networks on sequence
  modeling.
\newblock \emph{arXiv preprint arXiv:1412.3555}.

\bibitem[{Dosovitskiy et~al.(2020)Dosovitskiy, Beyer, Kolesnikov, Weissenborn,
  Zhai, Unterthiner, Dehghani, Minderer, Heigold, Gelly
  et~al.}]{dosovitskiy2020image}
Dosovitskiy, A.; Beyer, L.; Kolesnikov, A.; Weissenborn, D.; Zhai, X.;
  Unterthiner, T.; Dehghani, M.; Minderer, M.; Heigold, G.; Gelly, S.; et~al.
  2020.
\newblock An Image is Worth 16x16 Words: Transformers for Image Recognition at
  Scale.
\newblock In \emph{ICLR}.

\bibitem[{Garau et~al.(2022)Garau, Bisagno, Sambugaro, and
  Conci}]{garau2022interpretable}
Garau, N.; Bisagno, N.; Sambugaro, Z.; and Conci, N. 2022.
\newblock Interpretable part-whole hierarchies and conceptual-semantic
  relationships in neural networks.
\newblock In \emph{CVPR}, 13689--13698.

\bibitem[{Ge et~al.(2021)Ge, Xiao, Xu, Zheng, Karanam, Chen, Itti, and
  Wu}]{ge2021peek}
Ge, Y.; Xiao, Y.; Xu, Z.; Zheng, M.; Karanam, S.; Chen, T.; Itti, L.; and Wu,
  Z. 2021.
\newblock A peek into the reasoning of neural networks: Interpreting with
  structural visual concepts.
\newblock In \emph{CVPR}, 2195--2204.

\bibitem[{Ha and Eck(2017)}]{ha2017neural}
Ha, D.; and Eck, D. 2017.
\newblock A neural representation of sketch drawings.
\newblock \emph{arXiv preprint arXiv:1704.03477}.

\bibitem[{Ha and Eck(2018)}]{ha2018neural}
Ha, D.; and Eck, D. 2018.
\newblock A Neural Representation of Sketch Drawings.
\newblock In \emph{ICLR}.

\bibitem[{Khasahmadi et~al.(2019)Khasahmadi, Hassani, Moradi, Lee, and
  Morris}]{khasahmadi2019memory}
Khasahmadi, A.~H.; Hassani, K.; Moradi, P.; Lee, L.; and Morris, Q. 2019.
\newblock Memory-Based Graph Networks.
\newblock In \emph{ICLR}.

\bibitem[{Li et~al.(2019)Li, Muller, Thabet, and Ghanem}]{li2019deepgcns}
Li, G.; Muller, M.; Thabet, A.; and Ghanem, B. 2019.
\newblock Deepgcns: Can gcns go as deep as cnns?
\newblock In \emph{ICCV}, 9267--9276.

\bibitem[{Li et~al.(2018)Li, Pang, Song, Song, Xiang, Hospedales, and
  Zhang}]{li2018universal}
Li, K.; Pang, K.; Song, J.; Song, Y.-Z.; Xiang, T.; Hospedales, T.~M.; and
  Zhang, H. 2018.
\newblock Universal sketch perceptual grouping.
\newblock In \emph{ECCV}, 582--597.

\bibitem[{Li et~al.(2020)Li, Zou, Zheng, Su, Fu, and Tai}]{li2020sketch}
Li, L.; Zou, C.; Zheng, Y.; Su, Q.; Fu, H.; and Tai, C.-L. 2020.
\newblock Sketch-R2CNN: an RNN-rasterization-CNN architecture for vector sketch
  recognition.
\newblock \emph{IEEE TVCG}, 27(9): 3745--3754.

\bibitem[{Liu et~al.(2023)Liu, Li, Zhang, Xu, and Cao}]{liu2023prediction}
Liu, S.; Li, J.; Zhang, H.; Xu, L.; and Cao, X. 2023.
\newblock Prediction with Visual Evidence: Sketch Classification Explanation
  via Stroke-Level Attributions.
\newblock \emph{IEEE TIP}.

\bibitem[{Miller(2019)}]{miller2019explanation}
Miller, T. 2019.
\newblock Explanation in artificial intelligence: Insights from the social
  sciences.
\newblock \emph{Artificial intelligence}, 267: 1--38.

\bibitem[{Omeiza et~al.(2019)Omeiza, Speakman, Cintas, and
  Weldermariam}]{omeiza2019smooth}
Omeiza, D.; Speakman, S.; Cintas, C.; and Weldermariam, K. 2019.
\newblock Smooth Grad-CAM++: An Enhanced Inference Level Visualization
  Technique for Deep Convolutional Neural Network Models.
\newblock \emph{arXiv preprint arXiv:1908.01224}.

\bibitem[{Prabhu et~al.(2018)Prabhu, Batchu, Munagala, Gajawada, and
  Namboodiri}]{prabhu2018distribution}
Prabhu, A.; Batchu, V.; Munagala, S.~A.; Gajawada, R.; and Namboodiri, A. 2018.
\newblock Distribution-aware binarization of neural networks for sketch
  recognition.
\newblock In \emph{WACV}, 830--838.

\bibitem[{Qu et~al.(2023)Qu, Gryaditskaya, Li, Pang, Xiang, and
  Song}]{qu2023sketchxai}
Qu, Z.; Gryaditskaya, Y.; Li, K.; Pang, K.; Xiang, T.; and Song, Y.-Z. 2023.
\newblock SketchXAI: A First Look at Explainability for Human Sketches.
\newblock In \emph{CVPR}, 23327--23337.

\bibitem[{Ramaswamy et~al.(2020)}]{ramaswamy2020ablation}
Ramaswamy, H.~G.; et~al. 2020.
\newblock Ablation-cam: Visual explanations for deep convolutional network via
  gradient-free localization.
\newblock In \emph{WACV}, 983--991.

\bibitem[{Sarvadevabhatla and Kundu(2016)}]{sarvadevabhatla2016enabling}
Sarvadevabhatla, R.~K.; and Kundu, J. 2016.
\newblock Enabling my robot to play pictionary: Recurrent neural networks for
  sketch recognition.
\newblock In \emph{ACM MM}, 247--251.

\bibitem[{Selvaraju et~al.(2017)Selvaraju, Cogswell, Das, Vedantam, Parikh, and
  Batra}]{ramprasaath2017gradcam}
Selvaraju, R.~R.; Cogswell, M.; Das, A.; Vedantam, R.; Parikh, D.; and Batra,
  D. 2017.
\newblock Grad-CAM: Visual Explanations from Deep Networks via Gradient-Based
  Localization.
\newblock In \emph{ICCV}, 618--626.

\bibitem[{Shitole et~al.(2021)Shitole, Li, Kahng, Tadepalli, and
  Fern}]{shitole2021one}
Shitole, V.; Li, F.; Kahng, M.; Tadepalli, P.; and Fern, A. 2021.
\newblock One explanation is not enough: structured attention graphs for image
  classification.
\newblock volume~34, 11352--11363.

\bibitem[{Van~der Maaten and Hinton(2008)}]{van2008visualizing}
Van~der Maaten, L.; and Hinton, G. 2008.
\newblock Visualizing data using t-SNE.
\newblock \emph{JMLR}, 9(11).

\bibitem[{Van~Looveren and Klaise(2021)}]{van2021interpretable}
Van~Looveren, A.; and Klaise, J. 2021.
\newblock Interpretable counterfactual explanations guided by prototypes.
\newblock In \emph{Joint European Conference on Machine Learning and Knowledge
  Discovery in Databases}, 650--665.

\bibitem[{Wang et~al.(2019)Wang, Sun, Liu, Sarma, Bronstein, and
  Solomon}]{wang2019dynamic}
Wang, Y.; Sun, Y.; Liu, Z.; Sarma, S.~E.; Bronstein, M.~M.; and Solomon, J.~M.
  2019.
\newblock Dynamic graph cnn for learning on point clouds.
\newblock \emph{ACM TOG}, 38(5): 1--12.

\bibitem[{Wu et~al.(2018)Wu, Qi, Liu, and Yang}]{wu2018sketchsegnet}
Wu, X.; Qi, Y.; Liu, J.; and Yang, J. 2018.
\newblock Sketchsegnet: A rnn model for labeling sketch strokes.
\newblock In \emph{MLSP}, 1--6.

\bibitem[{Xu et~al.(2022)Xu, Hospedales, Yin, Song, Xiang, and
  Wang}]{xu2022deep}
Xu, P.; Hospedales, T.~M.; Yin, Q.; Song, Y.-Z.; Xiang, T.; and Wang, L. 2022.
\newblock Deep learning for free-hand sketch: A survey.
\newblock \emph{IEEE TPAMI}, 45(1): 285--312.

\bibitem[{Xu et~al.(2018)Xu, Huang, Yuan, Pang, Song, Xiang, Hospedales, Ma,
  and Guo}]{xu2018sketchmate}
Xu, P.; Huang, Y.; Yuan, T.; Pang, K.; Song, Y.-Z.; Xiang, T.; Hospedales,
  T.~M.; Ma, Z.; and Guo, J. 2018.
\newblock Sketchmate: Deep hashing for million-scale human sketch retrieval.
\newblock In \emph{CVPR}, 8090--8098.

\bibitem[{Xu, Joshi, and Bresson(2021)}]{xu2021multigraph}
Xu, P.; Joshi, C.~K.; and Bresson, X. 2021.
\newblock Multigraph transformer for free-hand sketch recognition.
\newblock \emph{IEEE TNNLS}, 33(10): 5150--5161.

\bibitem[{Yang et~al.(2021)Yang, Zhuang, Fu, Wei, Zhou, and
  Zheng}]{yang2021sketchgnn}
Yang, L.; Zhuang, J.; Fu, H.; Wei, X.; Zhou, K.; and Zheng, Y. 2021.
\newblock Sketchgnn: Semantic sketch segmentation with graph neural networks.
\newblock \emph{ACM TOG}, 40(3): 1--13.

\bibitem[{Yu et~al.(2017)Yu, Yang, Liu, Song, Xiang, and
  Hospedales}]{yu2017sketch}
Yu, Q.; Yang, Y.; Liu, F.; Song, Y.-Z.; Xiang, T.; and Hospedales, T.~M. 2017.
\newblock Sketch-a-net: A deep neural network that beats humans.
\newblock \emph{IJCV}, 122: 411--425.

\bibitem[{Yuan and Ji(2020)}]{yuan2020structpool}
Yuan, H.; and Ji, S. 2020.
\newblock Structpool: Structured graph pooling via conditional random fields.
\newblock In \emph{ICLR}.

\bibitem[{Zhang et~al.(2018)Zhang, Cao, Shi, Wu, and
  Zhu}]{zhang2018interpreting}
Zhang, Q.; Cao, R.; Shi, F.; Wu, Y.~N.; and Zhu, S.-C. 2018.
\newblock Interpreting CNN knowledge via an explanatory graph.
\newblock In \emph{AAAI}, 4454--4463.

\bibitem[{Zhang et~al.(2019)Zhang, Li, Liu, and Feng}]{zhang2019survey}
Zhang, X.; Li, X.; Liu, Y.; and Feng, F. 2019.
\newblock A survey on freehand sketch recognition and retrieval.
\newblock \emph{IMAVIS}, 89: 67--87.

\bibitem[{Zhu et~al.(2023)Zhu, Wang, Cheng, Wu, Li, and Zhang}]{zhu2023sketch}
Zhu, G.; Wang, S.; Cheng, Q.; Wu, K.; Li, H.; and Zhang, L. 2023.
\newblock Sketch Input Method Editor: A Comprehensive Dataset and Methodology
  for Systematic Input Recognition.
\newblock In \emph{ACM MM}.

\end{thebibliography}

\end{document}